\documentclass[11pt]{article}

\usepackage{microtype} 
\usepackage{booktabs}  
\usepackage{url}  
\usepackage{graphicx} 
\usepackage{multirow}
\usepackage{float}
\usepackage{amsmath}
\usepackage{amsthm}

\usepackage[english]{babel} 
\usepackage[final,shortpaper]{automl}
\usepackage{automl}

\usepackage[%
  backend=biber,
  style=authoryear-comp,
  sortcites=true,
  natbib=true,
  giveninits=true,
  maxcitenames=2,
  doi=false,
  url=true,
  isbn=false,
  dashed=false
]{biblatex}
\usepackage{colortbl}
\newcommand{\cdc}{\cellcolor[gray]{0.9}\textcolor{black}}
\newcommand{\cdcb}[1]{\cellcolor[gray]{0.9}\textcolor{black}{\textbf{#1}}}

\title{Tune My Adam, Please!}

\author[1]{\nameemail{Theodoros Athanasiadis}{athanast@informatik.uni-freiburg.de}}
\author[1]{\nameemail{Steven Adriaensen}{adriaens@cs.uni-freiburg.de}}
\author[1,2]{\nameemail{Samuel Müller}{sammuller@meta.com}}
\author[1,3,4]{\nameemail{Frank Hutter}{fh@cs.uni-freiburg.de}}

\affil[1]{University of Freiburg, Freiburg, Germany}
\affil[2]{Meta, New York City, USA}
\affil[3]{ELLIS Institute Tübingen, Tübingen, Germany}
\affil[4]{Prior Labs, Freiburg, Germany}
\hypersetup{%
  pdfauthor={Theodoros Athanasiadis, Steven Adriaensen, Samuel Müller}, 
  pdftitle={Tune My Adam, Please!},
  pdfsubject={Learning curve augmentation and surrogate modeling for tuning Adam optimizer with Freeze-Thaw BO.},
  pdfkeywords={adam,pfn,learning,curve,extrapolation,surrogate,bayesian,optimization,freeze,thaw,BO,}
}

\usepackage{amssymb}
\usepackage{pifont}
\newcommand{\cmark}{\ding{51}}%

\begin{document}

\maketitle

\begin{abstract}
    The Adam optimizer remains one of the most widely used optimizers in deep learning, and effectively tuning its hyperparameters is key to optimizing performance. However, tuning can be tedious and costly. Freeze-thaw Bayesian Optimization (BO) is a recent promising approach for low-budget hyperparameter tuning, but is limited by generic surrogates without prior knowledge of how hyperparameters affect learning. We propose \texttt{Adam-PFN},  a new surrogate model for Freeze-thaw BO of Adam's hyperparameters, pre-trained on learning curves from TaskSet, together with a new learning curve augmentation method, CDF-augment, which artificially increases the number of available training examples. Our approach improves both learning curve extrapolation and accelerates hyperparameter optimization on TaskSet evaluation tasks, with strong performance on out-of-distribution (OOD) tasks. 
\end{abstract}


\section{Introduction}
Hyperparameter Optimization (HPO) can be costly and time-consuming, especially in the age of LLMs. Although Bayesian Optimization (BO) is a sample-efficient alternative, it still requires training a model in full to get a new evaluation point for the BO loop. 

Freeze-thaw BO (FT-BO) \autocite{swersky_ft_2014} is a more efficient alternative that allocates resources to the most prominent configurations one step (e.g., one epoch) at a time. Additionally, rather than disregarding a configuration, FT-BO keeps it frozen in memory and can decide whether to thaw that configuration or start a new one.

Recently, \cite{ifbo} introduced \texttt{ifBO}, a new, state-of-the-art framework for FT-BO. It consists of \texttt{FT-PFN}, a PFN \autocite{pfn} based surrogate that performs Bayesian learning curve extrapolation, trained on synthetic hyperparameter (HP) configurations and learning curve combinations. However, \texttt{ifBO} attempts to fit every HPO case with one surrogate model.

Our work introduces a drop-in replacement surrogate model for \texttt{ifBO}, \texttt{Adam-PFN}, specialized to tune the HPs of the Adam optimizer \autocite{kingma_adam}. We also introduce a learning curve augmentation method that allows us to train on real data from TaskSet \autocite{taskset}, a collection of learning curves optimized using Adam, instead of sampling them from a synthetic prior.

\section{Adam-PFN}
\subsection{Real Data: TaskSet}
   TaskSet offers a collection of 1162 learning curves on a variety of tasks, including language modeling on words, subwords, and characters with RNNs, text classification with RNNs, CNNs trained on image data, and MLPs, all optimized with Adam. Each task contains the learning curves of the same 1000 randomly sampled HP configurations, including the learning rate, $\beta_1$, $\beta_2$, $\epsilon$, two parameters that control $L_1$ and $L_2$ regularization, and two parameters that control the learning rate decay schedule. All hyperparameter values are uniformly sampled on a log scale.  
    Table \ref{tab:taskset_hp_ranges} in Appendix \ref{appendix:taskset_hyperparameter_bounds} gives a more detailed overview of the hyperparameters' lower and upper bounds.

    \subsection{Augmentation Method: CDF-augment}
    To generate an even more diverse training set, we introduce CDF-augment, a local learning curve augmentation method that uses the cumulative distribution function (CDF) of the Beta distribution. 

    To augment a learning curve, we first sample Beta's mode $\mu$ uniformly in [0, 1] and concentration $\kappa$ uniformly in [2, 5], and calculate the CDF. We then forward-propagate the learning curve $y$ through the CDF as follows
    \begin{equation}
        \label{eqn:cdf_y}
        y' = F_{Beta}(y;\mu, \kappa)
    \end{equation}

\begin{figure} 
        \centering
        \begin{subfigure}[b]{0.35\textwidth}
            \centering
            \includegraphics[width=\textwidth]{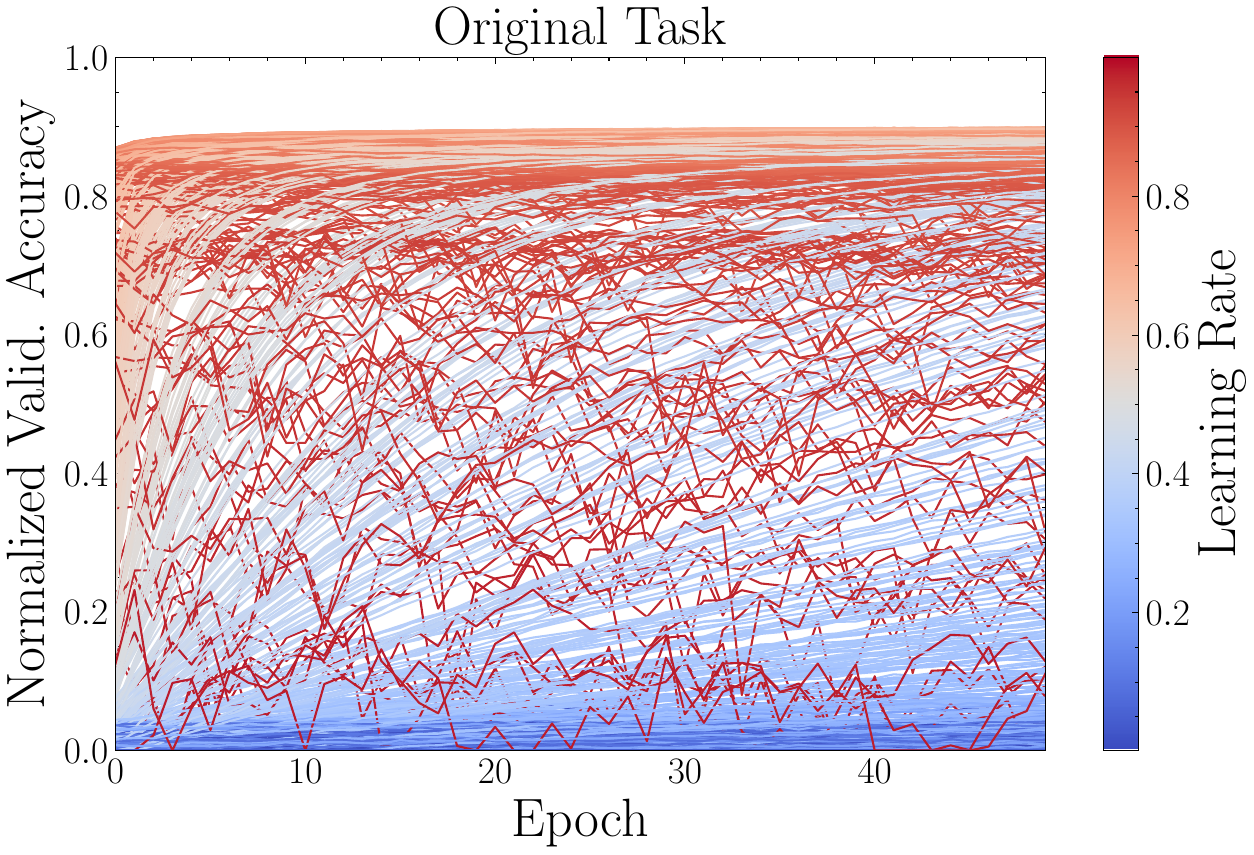}
            \caption{}
        \end{subfigure}
        \begin{subfigure}[b]{0.35\textwidth}
            \centering
            \includegraphics[width=\textwidth]{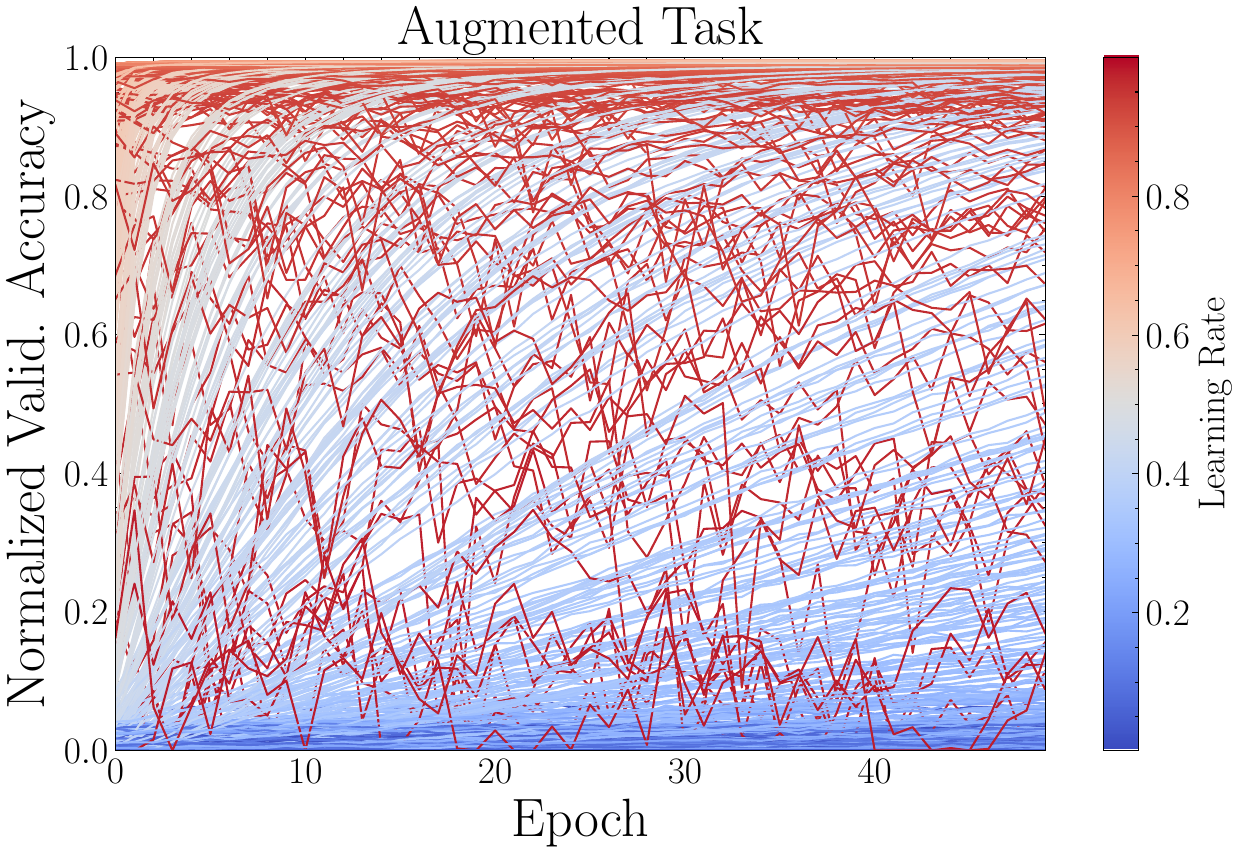}
            \caption{}
        \end{subfigure}
        \caption{(a) A selection of learning curves from TaskSet, the color of each curve is based on the learning rate value. (b) The same set of learning curves augmented with CDF-augment. CDF-augment non-linearly transformed "task hardness" by making the task easier.
        }
        \label{fig:cdf_augment}
    \end{figure}

    Using this augmentation, we increase the learning curve diversity by non-linearly transforming task-hardness. Due to the nature of the CDF, the rank (ordering) of learning curves is preserved. Figure \ref{fig:cdf_augment} shows the result of applying CDF-augment on a set of learning curves.
    
    \subsection{Surrogate Model Training}
    Our surrogate model is a PFN, trained on augmented data from real learning curves of the Adam optimizer, similar to the data it will encounter in practice. PFNs are based on the transformer architecture \autocite{vaswani_attention_2017}, and leverage in-context learning to perform Bayesian learning curve extrapolation in a single forward pass.

    We follow the same training pipeline for our surrogate model as \texttt{FT-PFN}, the only difference is that we do not sample the learning curves with and their HP configurations from a synthetic prior, rather, we train on real curves from TaskSet, which we augment using CDF-augment. We include more implementation details in Appendix \ref{appendix:implementation_details}.
    
    Starting with a set of learning curves randomly sampled from 878 tasks and 1000 HP configurations per task, we augment and then split each curve into context and query points. We train our model to extrapolate the performance of a curve at a query point, given the context and HP configuration. More details on the training tasks can be found in Appendix \ref{appendix:list_of_taskset_training_tasks}.

\section{Experiments}
    \subsection{Evaluation Procedure}
    Our evaluation set consists of 12 NLP tasks from TaskSet, which were not used during training and are also used in prior work \autocite{kadra_dpl_2023}, \autocite{wistuba_dyhpo_2023}, \autocite{ifbo}. A list of the evaluation tasks is included in Appendix \ref{appendix:evaluation_tasks}.
    
    To evaluate the quality and runtime complexity of predictions, we compare our approach against \texttt{DyHPO} \autocite{wistuba_dyhpo_2023}, \texttt{DPL} \autocite{kadra_dpl_2023}, \texttt{FT-PFN} \autocite{ifbo}, and a Uniform Predictor that outputs a uniform distribution over the [0, 1] range for Log-likelihood and a constant value of 0.5 for MSE.

    We also evaluated the HPO performance of our surrogate as part of \texttt{ifBO} by replacing \texttt{FT-PFN} while keeping the proposed acquisition function. For HPO performance, we additionally compare against \texttt{HyperBand} \autocite{li_hyperband_2018}, \texttt{ASHA} \autocite{asha}, \texttt{Freeze-thaw with GPs} \autocite{swersky_ft_2014}, and \texttt{Random Search} \autocite{random_search}. We include additional details on the baselines and discuss how they relate to our work in Appendix \ref{appendix:baselines}.

    We also included two variants of Adam-PFN, "No Aug" and "Mixup". \texttt{Adam-PFN (No aug.)} was trained on the initial TaskSet tasks, without any learning curve augmentation, while \texttt{Adam-PFN (Mixup)} was trained using the recently introduced Mixup learning curve augmentation method \autocite{lee_cmbo_2024}. Further details on Mixup are included in Appendix \ref{appendix:mixup}.

\subsection{Learning Curve Extrapolation Results}
    Table \ref{table:Taskset8p_main} presents the learning curve extrapolation results. \texttt{Adam-PFN (CDF)} outperforms all other baselines, both in terms of Log-likelihood (LL) and MSE. As expected, the inference time of all PFN approaches is significantly lower than that of DPL and DyHPO, since they are trained offline.

    \begin{table}
        \centering
        \scalebox{0.75}{
        \begin{tabular}{l ccc ccc ccc}
            \toprule
            \multirow{2}{*}{\textbf{Algorithm}} &
            \multicolumn{3}{c}{\textbf{Context 400}} &
            \multicolumn{3}{c}{\textbf{Context 1000}} &
            \multicolumn{3}{c}{\textbf{Context 1600}} \\
            \cmidrule(lr){2-4} \cmidrule(lr){5-7} \cmidrule(lr){8-10}
            & LL & MSE & Time & LL & MSE & Time & LL & MSE & Time \\
            \midrule
            Adam-PFN (CDF)     & \textbf{5.326} & \textbf{0.00054} & 0.697 & \textbf{5.422} & \textbf{0.00046} & 2.151 & \textbf{5.441} & \textbf{0.00043} & 4.014 \\
            Adam-PFN (Mixup)   & 4.916 & 0.00080 & 0.677 & 4.986 & 0.00067 & 2.131 & 5.022 & 0.00061 & 3.972 \\
            Adam-PFN (No aug.) & 4.947 & 0.00062 & 0.671 & 5.042 & 0.00054 & 2.090 & 5.066 & 0.00052 & 3.993 \\
            FT-PFN            & 3.440 & 0.00184 & 0.693 & 3.473 & 0.00171 & 2.103 & 3.443 & 0.00164 & 4.017 \\
            DPL               & -29.752 & 0.00191 & 17.790 & -22.192 & 0.00175 & 38.855 & -16.154 & 0.00164 & 63.748 \\
            DyHPO             & -0.494 & 0.00336 & 25.662 & -0.397 & 0.00317 & 113.217 & -0.383 & 0.00317 & 264.284 \\
            Uniform           & -6.908 & 0.22881 & N/A & -6.908 & 0.22852 & 0.000 & -6.908 & 0.22896 & N/A \\
            \bottomrule
        \end{tabular}
        }
        \caption{Comparison of \texttt{Adam-PFN(CDF)}, \texttt{Adam-PFN(Mixup)}, and \texttt{Adam-PFN (No Aug.)} against the baselines for different context sizes. Values are the median across the evaluation tasks.}
        \label{table:Taskset8p_main}
    \end{table}

    \subsection{HPO Results}
\begin{figure}
            \centering
            \includegraphics[width=0.8\linewidth]{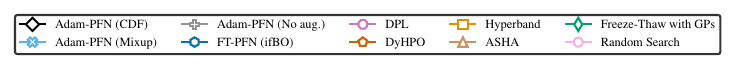}
    
            \begin{subfigure}{0.37\linewidth}
                \centering
                \includegraphics[width=\linewidth]{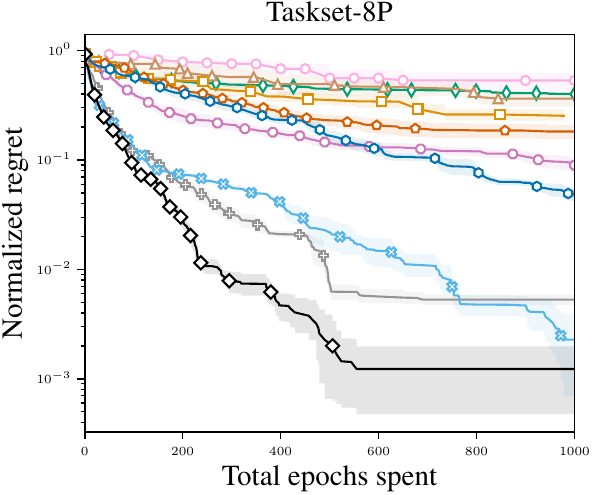}
                \caption{Normalized regret}
            \end{subfigure}
            \begin{subfigure}{0.37\linewidth}
                \centering
                \includegraphics[width=\linewidth]{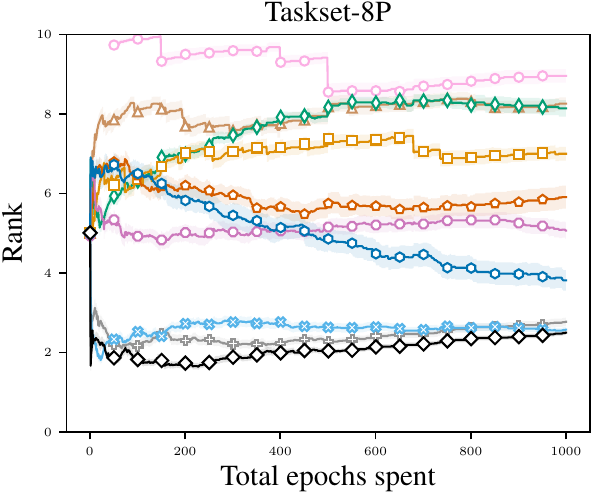}
                \caption{Average rank}
            \end{subfigure}
            \caption{HPO results of normalized regret and average rank of \texttt{Adam-PFN (CDF)} against the baselines. The results are the mean across the 12 evaluation tasks for 5 random seeds.}
            \label{fig:8p_main_results}
        \end{figure}

    Figure \ref{fig:8p_main_results} shows the aggregated normalized regret and the average rank of each algorithm. \texttt{Adam-PFN (CDF)} outperforms all baselines both in terms of normalized regret and average rank. The other two alternatives, \texttt{Adam-PFN (Mixup)} and \texttt{Adam-PFN (No. aug)}, perform similarly well. All variants are also sample-efficient, achieving the same normalized regret at approximately epoch 150, as \texttt{FT-PFN} does at approximately epoch 750. 

\subsection{Evaluation In The Wild: Out-Of-Distribution Tasks}
    We also evaluated the HPO performance of all \texttt{Adam-PFN} variants on some real-world tasks from the publicly available PyTorch Examples\footnote{https://github.com/pytorch/examples} repository. We compare our approach with \texttt{FT-PFN}, HyperBand, ASHA, and Random Search.

\begin{figure}
                \centering
                \includegraphics[width=0.8\linewidth]{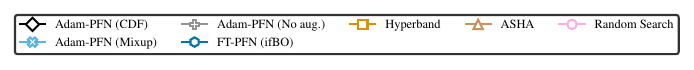}
        
                \begin{subfigure}{0.37\linewidth}
                    \centering
                    \includegraphics[width=\linewidth]{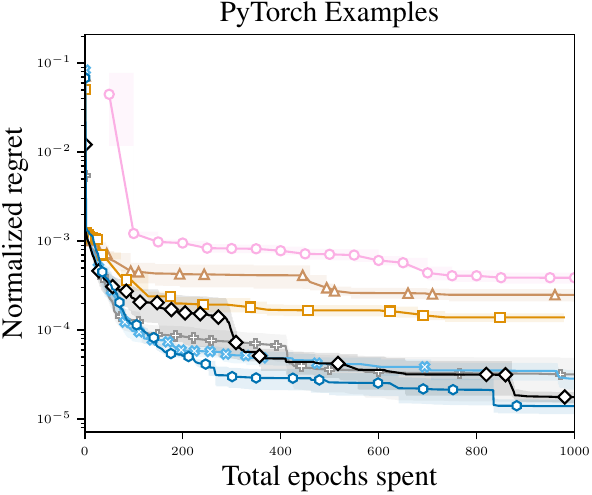}
                    \caption{Normalized regret}
                \end{subfigure}
                \begin{subfigure}{0.37\linewidth}
                    \centering
                    \includegraphics[width=\linewidth]{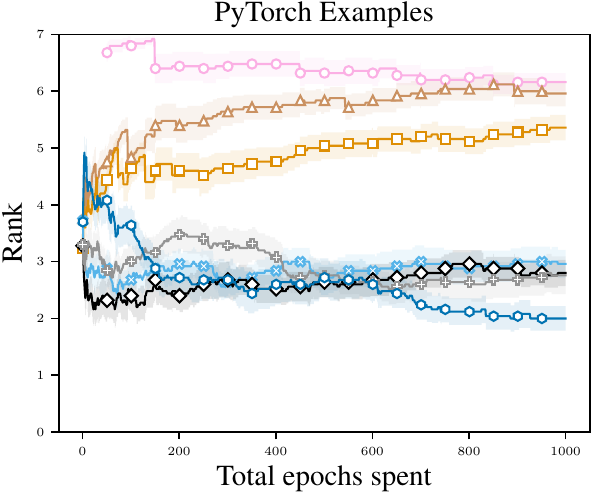}
                    \caption{Average rank}
                \end{subfigure}
                \caption{HPO results of normalized regret and average rank of \texttt{Adam-PFN (CDF)} against the baselines on OOD tasks. The results are the mean across tasks for 5 random seeds.}
                \label{fig:pytorch_examples_hpo_results}
    \end{figure}

    The results are presented in Figure \ref{fig:pytorch_examples_hpo_results}. Focusing on subfigure (b), which shows the average ranks, we observe that \texttt{Adam-PFN (CDF)} is the best-performing alternative early on, on average. As the budget increases, \texttt{FT-PFN} catches up and eventually surpasses our model. An approach that warm-starts the HPO process with \texttt{Adam-PFN} and then switches to \texttt{FT-PFN} might be optimal for these tasks, but we leave that as future work. We include more details and results in Appendix \ref{appendix:ood_results_per_task}.
    
\section{Limitations \& Future Work}
\label{section:limitations_and_future_work}
Our preliminary results highlight the importance of specialization and the use of prior knowledge when it comes to HPO. Despite that, we acknowledge that our surrogate model is trained and tested on a fixed search space with a pre-defined number of HPs. Tuning different sets of HPs needs to be further explored. One possible approach would be to set unused HPs to default values during training and testing. 

Unlike Mixup, we only consider augmentations in learning curve space. While we explored HP augmentation, we surprisingly found that augmenting in the parameter space hurts performance, whereas Mixup's results improve when we remove HP augmentation (refer to Appendix \ref{appendix:hp_augmentation}). We believe that our results could be further improved with the introduction of a new HP augmentation method.

Finally, especially based on the results on OOD tasks, we believe it is worth exploring how our prior could be mixed with that of \texttt{FT-PFN}. One possibility would be to train a model from scratch with a prior that is a mixture of the two. Another would be to fine-tune \texttt{FT-PFN} on learning curves from our prior.

\begin{acknowledgements}
  Frank Hutter is a Hector Endowed Fellow at the ELLIS Institute Tubingen. All authors acknowledge funding by the state of Baden-Wurttemberg through bwHPC, the German Research Foundation (DFG) through grant number 417962828, and the European Union (via ERC Consolidator Grant Deep Learning 2.0, grant no. 101045765). Views and opinions expressed are, however, those of the author(s) only and do not necessarily reflect those of the European Union or the European Research Council. Neither the European Union nor the granting authority can be held responsible for them.

  \begin{center}\includegraphics[width=0.3\textwidth]{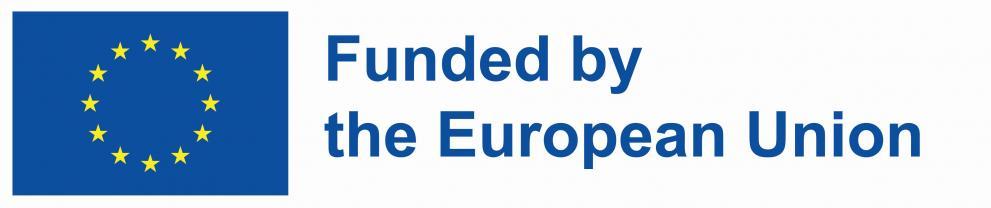}\end{center}
\end{acknowledgements}



\printbibliography


\newpage

\appendix

\section{TaskSet Hyperparameter Bounds}
\label{appendix:taskset_hyperparameter_bounds}
    Table \ref{tab:taskset_hp_ranges} provides the sampling bounds for TaskSet hyperparameters. We normalize all hyperparameters to lie on the [0, 1] range using those bounds before we feed them to the model for training and inference.

    \begin{table}[ht!]
        \centering
        \begin{tabular}{@{\hskip 10pt} l c c c @{\hskip 10pt}}
            \toprule
             \textbf{HP } & \textbf{Log} & \textbf{Lower Bound} & \textbf{Upper Bound} \\
             \midrule
             Learning Rate  & \cmark    & $1e^{-8}$ & 10        \\
             $\beta_1$\footnotemark      & \cmark    & $1e^{-4}$ & 1         \\
             $\beta_2$\footnotemark[\value{footnote}]     & \cmark    & $1e^{-6}$ & 1         \\
             Linear Decay   & \cmark    & $1e^{-7}$ & $0.0001$ \\
             Expon. Decay   & \cmark    & $1e^{-6}$ & $0.001$ \\
             $\epsilon$     & \cmark    & $1e^{-10}$ & $1000$ \\
             $\ell_1$       & \cmark    & $1e^{-8}$ & 1 \\
             $\ell_2$       & \cmark    & $1e^{-8}$ & 1 \\
            \bottomrule
        \end{tabular}
        \caption{Sampling strategies for TaskSet Hyperparameters.}
        \label{tab:taskset_hp_ranges}
    \end{table}

    \footnotetext{$\beta_1$ and $\beta_2$ are parametrized as $1-x$ and then sampled.}

\section{Adam-PFN Implementation Details}
\label{appendix:implementation_details}
    Following \cite{ifbo}, we used a transformer \autocite{vaswani_attention_2017} with 6 layers and 4 attention heads, an embedding size of 512, and a hidden size of 1024. During training, we used a batch size of 25 and the Adam \autocite{kingma_adam} optimizer with an initial learning rate of 0.0001. We scheduled the learning rate using cosine annealing \autocite{loshchilov2017sgdrstochasticgradientdescent} with a linear warm-up schedule for the first 200 epochs. We train on a total of 2.0M learning curves. A list of the sampled tasks we used during training is presented in Appendix \ref{appendix:list_of_taskset_training_tasks}.

\section{List of TaskSet Training Tasks}
\label{appendix:list_of_taskset_training_tasks}

TaskSet provides both hand-designed tasks designed by experts, and sampled tasks, where the task parameters such as dataset, network architecture, activation functions, etc., are randomly sampled. There are far more sampled tasks in TaskSet, since they are easier to create. In our work, we exclusively train on sampled tasks. This was a design decision to highlight the fact that the learning curves we use during training do not have to come from carefully curated tasks. Instead, randomly sampling a task's parameters and optimizing it for a number of epochs suffices.

To facilitate reproducibility, we provide a detailed list of each task we used during training.

\begin{itemize}[leftmargin=*,noitemsep]
    \item \texttt{word\_rnn\_language\_model\_family} seeds: 0, 1, 3, 4, 5, 8, 9, 10, 11, 13, 14, 15, 16, 17, 19, 21, 22, 23, 24, 25, 26, 29, 30, 31, 33, 34, 37, 39, 41, 42, 43, 45, 46, 47, 48, 51, 52, 53, 54, 55, 57, 58, 59, 61, 63, 64, 65, 66, 67, 69, 72, 73, 74, 76, 78, 81, 82, 84, 85, 86, 87, 88, 89, 91, 93, 94, 97, 98, 99

    \item \texttt{char\_rnn\_language\_model\_family} seeds: 0, 1, 2, 3, 4, 5, 6, 7, 8, 9, 10, 11, 12, 13, 14, 15, 16, 17, 18, 19, 20, 21, 22, 23, 24, 25, 26, 27, 28, 29, 30, 31, 33, 34, 36, 37, 38, 39, 40, 41, 42, 43, 45, 46, 47, 48, 49, 51, 52, 53, 54, 55, 56, 57, 58, 59, 60, 61, 62, 63, 64, 65, 66, 67, 68, 69, 70, 71, 72, 73, 74, 75, 76, 78, 79, 80, 81, 82, 83, 84, 85, 86, 87, 88, 89, 90, 91, 92, 93, 94, 95, 96, 97, 98, 99
    
    \item \texttt{conv\_fc\_family} seeds: 0, 1, 2, 3, 4, 5, 6, 7, 8, 9, 10, 11, 12, 13, 14, 15, 16, 17, 18, 19, 20, 21, 22, 23, 24, 26, 27, 28, 30, 31, 32, 34, 35, 36, 37, 38, 39, 40, 41, 42, 44, 45, 46, 47, 48, 49, 50, 51, 52, 53, 54, 55, 57, 58, 59, 60, 61, 62, 63, 64, 65, 67, 68, 69, 70, 72, 73, 74, 75, 76, 77, 78, 79, 80, 81, 82, 83, 84, 85, 86, 87, 88, 89, 90, 91, 92, 93, 94, 95, 96, 97, 98, 99
    
    \item \texttt{conv\_pooling\_family} seeds: 0, 1, 2, 3, 4, 5, 6, 7, 8, 9, 10, 11, 12, 13, 14, 15, 16, 17, 18, 19, 20, 21, 22, 23, 24, 25, 26, 27, 28, 30, 31, 32, 34, 35, 36, 37, 38, 39, 40, 41, 44, 45, 46, 47, 48, 49, 50, 51, 52, 53, 54, 55, 57, 58, 59, 60, 61, 62, 63, 64, 65, 67, 68, 69, 70, 72, 73, 74, 75, 76, 77, 78, 79, 80, 81, 82, 83, 84, 85, 86, 87, 88, 89, 90, 91, 92, 93, 94, 95, 96, 97, 98, 99
    
    \item \texttt{losg\_tasks\_family} seeds: 0, 1, 2, 5, 6, 7, 8, 9, 10, 11, 12, 13, 15, 16, 17, 18, 20, 21, 22, 23, 24, 25, 26, 27, 28, 29, 30, 31, 34, 35, 36, 37, 38, 39, 40, 41, 42, 43, 44, 45, 48, 50, 51, 52, 53, 54, 56, 57, 58, 59, 61, 62, 63, 65, 66, 67, 68, 69, 70, 71, 72, 73, 74, 75, 76, 78, 80, 81, 82, 83, 84, 85, 87, 88, 89, 90, 91, 92, 93, 94, 95, 97, 98, 99
    
    \item \texttt{maf\_family} seeds: 3, 5, 7, 14, 17, 20, 23, 27, 40, 41, 44, 46, 47, 48, 60, 61, 64, 66, 68, 69, 70, 76, 78, 83, 84, 89, 94
    
    \item \texttt{mlp\_ae\_family} seeds: 0, 1, 2, 3, 4, 5, 6, 7, 8, 9, 10, 11, 12, 13, 14, 15, 16, 17, 18, 19, 20, 21, 22, 23, 24, 25, 26, 27, 28, 29, 30, 31, 32, 33, 34, 35, 36, 37, 38, 39, 40, 41, 42, 43, 44, 45, 46, 48, 49, 50, 51, 52, 53, 54, 55, 56, 57, 58, 59, 60, 61, 62, 64, 65, 66, 67, 68, 69, 70, 71, 72, 73, 74, 75, 76, 77, 78, 79, 80, 81, 82, 83, 84, 85, 87, 88, 89, 90, 91, 92, 93, 94, 95, 96, 97, 98, 99
    
    \item \texttt{mlp\_family} seeds: 0, 1, 2, 3, 4, 5, 6, 7, 8, 9, 10, 11, 12, 13, 14, 15, 16, 17, 18, 19, 20, 21, 22, 23, 24, 25, 26, 27, 28, 29, 30, 31, 32, 33, 34, 35, 36, 37, 38, 39, 40, 41, 42, 43, 44, 45, 46, 47, 48, 49, 50, 51, 52, 53, 54, 55, 56, 57, 58, 59, 60, 61, 62, 63, 64, 65, 66, 67, 68, 69, 70, 71, 72, 73, 74, 75, 76, 77, 78, 80, 81, 82, 83, 84, 85, 86, 87, 88, 89, 90, 91, 92, 93, 94, 95, 96, 97, 98, 99
    
    \item \texttt{mlp\_vae\_family} seeds: 1, 3, 4, 7, 9, 10, 12, 14, 16, 18, 19, 20, 21, 22, 23, 24, 25, 30, 32, 33, 34, 35, 36, 37, 41, 42, 45, 50, 52, 54, 56, 57, 59, 60, 61, 62, 66, 67, 69, 71, 76, 77, 78, 79, 80, 81, 82, 83, 84, 86, 87, 90, 92, 94, 97
    
    \item \texttt{nvp\_family} seeds: 5, 7, 14, 19, 27, 28, 32, 35, 36, 42, 47, 49, 55, 65, 70, 73, 74, 77, 88, 89, 92, 93, 95
    
    \item \texttt{quadratic\_family} seeds: 0, 1, 5, 7, 8, 9, 10, 11, 12, 13, 14, 18, 19, 21, 23, 24, 25, 28, 31, 32, 34, 35, 36, 38, 42, 43, 44, 48, 53, 62, 63, 64, 66, 68, 69, 70, 72, 73, 75, 78, 81, 82, 83, 85, 89, 91, 93, 97
    
    \item \texttt{rnn\_text\_classification\_family} seeds: 0, 1, 2, 3, 4, 5, 6, 7, 8, 9, 10, 11, 12, 13, 14, 15, 16, 17, 18, 19, 20, 21, 22, 23, 24, 25, 26, 27, 28, 30, 31, 32, 33, 34, 35, 36, 37, 39, 40, 41, 42, 43, 44, 45, 46, 47, 49, 50, 51, 52, 53, 54, 55, 56, 57, 58, 59, 60, 61, 62, 63, 64, 65, 66, 67, 68, 69, 70, 71, 73, 74, 75, 76, 77, 78, 79, 80, 81, 82, 83, 84, 85, 86, 87, 88, 89, 90, 91, 93, 94, 95, 96, 97, 98, 99
\end{itemize}

\section{List of TaskSet Evaluation Tasks}
\label{appendix:evaluation_tasks}
We evaluated our pipeline on the following 12 text classification tasks, which are also used in prior work \autocite{ifbo}, \autocite{wistuba_dyhpo_2023}, \autocite{kadra_dpl_2023}:

    \begin{itemize}[itemsep=0pt, parsep=0pt]
        \item FixedTextRNNClassification\_imdb\_patch128\_LSTM128\_avg\_bs64
        \item FixedTextRNNClassification\_imdb\_patch128\_LSTM128\_bs64
        \item FixedTextRNNClassification\_imdb\_patch128\_LSTM128\_embed128\_bs64
        \item FixedTextRNNClassification\_imdb\_patch32\_GRU128\_bs128
        \item FixedTextRNNClassification\_imdb\_patch32\_GRU64\_avg\_bs128
        \item FixedTextRNNClassification\_imdb\_patch32\_IRNN64\_relu\_avg\_bs128
        \item FixedTextRNNClassification\_imdb\_patch32\_IRNN64\_relu\_last\_bs128
        \item FixedTextRNNClassification\_imdb\_patch32\_LSTM128\_E128\_bs128
        \item FixedTextRNNClassification\_imdb\_patch32\_LSTM128\_bs128
        \item FixedTextRNNClassification\_imdb\_patch32\_VRNN128\_tanh\_bs128
        \item FixedTextRNNClassification\_imdb\_patch32\_VRNN64\_relu\_avg\_bs128
        \item FixedTextRNNClassification\_imdb\_patch32\_VRNN64\_tanh\_avg\_bs128
    \end{itemize}
    \vspace{\baselineskip}

\section{Baselines}
\label{appendix:baselines}
\begin{enumerate}[wide, labelindent=0pt, itemsep=0em]
    \item[\textbf{DyHPO}] \autocite{wistuba_dyhpo_2023} includes a learned deep kernel GP as the surrogate model, combined with a generalization of the Expected Improvement (EI) acquisition function for a multi-fidelity setup. The candidate configuration suggested by the acquisition function is evaluated only for a number of fidelity steps. During evaluation we used the NePS \autocite{Stoll_Neural_Pipeline_Search_2023} framework implementation provided by \cite{ifbo}, which follows the implementation details and code by \cite{wistuba_dyhpo_2023} from their publicly available repository at \href{https://github.com/machinelearningnuremberg/DyHPO}{\texttt{https://github.com/machinelearningnuremberg/DyHPO}}.
    
    \item[\textbf{DPL}] \autocite{kadra_dpl_2023} is a framework that performs Bayesian Optimization with learning curve extrapolation under the assumption that learning curves follow power law functions. Using an ensemble of neural networks as a surrogate model, and Expected Improvement (EI) at the maximum budget as the acquisition function, the candidate configurations are again trained for a number of epochs before the next BO iteration. During evaluation, we used the NePS framework implementation provided by \cite{ifbo}, which follows the implementation details and code by \cite{kadra_dpl_2023} from their publicly available repository at \href{https://github.com/machinelearningnuremberg/DPL}{\texttt{https://github.com/machinelearningnuremberg/DPL}}.
    
    \item[\textbf{FT-PFN}] is the surrogate model introduced by \cite{ifbo}, it is a PFN \autocite{pfn} that performs Bayesian learning curve extrapolation, trained on synthetic datasets created by 4 weighted basis functions with additive Gaussian noise. During evaluation, we used the implementation suggested by \cite{ifbo} and their publicly available surrogate model at \href{https://github.com/automl/ifBO}{\texttt{https://github.com/automl/ifBO}}.

    \item[\textbf{Uniform Predictor}] is a simple baseline that outputs a uniform distribution over the [0, 1] range for the log-likelihood and a constant value of 0.5 for the MSE point estimate.

    \item[\textbf{Freeze-Thaw with GPs}] \autocite{swersky_ft_2014} Freeze-Thaw using GPs as the surrogate model. During evaluation, we follow the same implementation used by \cite{ifbo}.
        
    \item[\textbf{Hyperband}] \autocite{li_hyperband_2018} is a Multi-fidelity algorithm which uses different Successive Halving \autocite{successive_halving} brackets with different definitions of lower fidelity per bracket. During evaluation, we used the default NePS implementation with $\eta=3$, which leads to 3 Succesive Halving brackets. The minimum budget was set to 1 and the maximum to 50. The cutoff fidelities for the first SH bracket were [1, 5, 16, 50].
    
    \item[\textbf{ASHA}] \autocite{asha} the asynchronous version of Successive Halving. During evaluation, we used the default NePS implementation, which uses $\eta=3$. The minimum budget was set to 1, and the maximum to 50.
    
    \item[\textbf{Random Search}] \autocite{random_search} an uninformed global optimization algorithm that samples hyperparameter configurations uniformly at random. During evaluation, we used the default NePS implementation. We allowed a total number of 1000 optimization steps, corresponding to 20 full random search evaluations in TaskSet tasks.
\end{enumerate}

\section{CMBO \& Mixup}
\label{appendix:mixup}
    CMBO \autocite{lee_cmbo_2024} is a recently introduced FT-BO framework to which our work is closely related. \texttt{CMBO} is a Cost-Sensitive Multi-fidelity BO framework that introduces a new utility function, aiming to balance a user-specific tradeoff between the expected performance improvement gain during an HPO run, and the computational resources cost needed to achieve that gain. 

    Similar to \cite{ifbo}, CMBO uses a PFN surrogate model, but unlike \texttt{ifBO}, the surrogate is trained on tasks from TaskSet, using Mixup, a learning curve augmentation method based on Mixup for images \autocite{zhang2018_mixup}.

    \texttt{CMBO} differs from \texttt{ifBO} in several important ways. Unlike \texttt{ifBO}, the surrogate model in \texttt{CMBO} is meta-learned using augmented learning curves from real tasks. A second key difference lies in the stopping criterion. \texttt{CMBO} uses a user-defined utility function to determine whether to continue the optimization, whereas \texttt{ifBO} runs until the full optimization budget is exhausted. Lastly, \texttt{CMBO} is not yet publicly available.

    Our approach also differs from \texttt{CMBO} in several ways. Our surrogate is trained on augmented curves from 878 tasks, compared to only 21 used in \texttt{CMBO}. We evaluated our model on both hand-designed NLP tasks and real-world tasks. Additionally, we investigate an alternative to Mixup that does not include the HP augmentation component (see also Appendix~\ref{appendix:hp_augmentation}). Lastly, \texttt{CMBO} is a new FT-BO framework, while our work builds upon the pre-existing open source \texttt{ifBO} and offers a drop-in replacement for its surrogate model. 
    
    For completeness, this appendix section describes Mixup, though more details can be found in the original paper by \cite{lee_cmbo_2024}\footnote{All the equations in this section are from \cite{lee_cmbo_2024}}.

    Mixup consists of two discrete augmentation steps. The first step involves augmenting the same hyperparameter configurations across tasks, and the second step involves augmenting across hyperparameter configurations within the same task.

    Assume $M$ tasks and $LC_m$ a vector of learning curve - hyperparameter configuration combinations for task $m \in M$, 
    \begin{equation}
        LC_m = [LC_{m,1}, LC_{m,2}, LC_{m,3}, ..., LC_{m,n}]
    \end{equation} 
    
    During task mixup, \cite{lee_cmbo_2024} sample a scalar value uniformly at random $\lambda_1 \sim Unif(0,1)$ and perform learning curve augmentation as follows:
    \begin{equation}
    \label{eqn::task_mixup_equation}
        LC'_m = \lambda_1LC_m + (1-\lambda_1)LC_{m'} \text{ for all }m,m' \in M 
    \end{equation}

    This step produces a new learning curve for each hyperparameter configuration as a linear interpolation between the learning curves of tasks $m$ and $m'$. 

    The second step involves mixing hyperparameter configurations and learning curves within the same task. Assume we have $N$ hyperparameter configuration - learning curve pairs denoted as $(x, y)$  per task. During hyperparameter mixup, \cite{lee_cmbo_2024} sample a scalar value uniformly at random $\lambda_2 \sim Unif(0, 1)$ and perform the augmentation as follows:
    \begin{equation}
    \label{eqn::hp_mixup_equation}
        (x'', y'') = \lambda_2 (x, y) + (1-\lambda_2)(x', y')
    \end{equation}

\begin{figure} 
        \centering
        \begin{subfigure}[b]{0.35\textwidth}
            \centering
            \includegraphics[width=\textwidth]{images/paper/cdf_original_task_no_hp.pdf}
            \caption{}
        \end{subfigure}
        \begin{subfigure}[b]{0.35\textwidth}
            \centering
            \includegraphics[width=\textwidth]{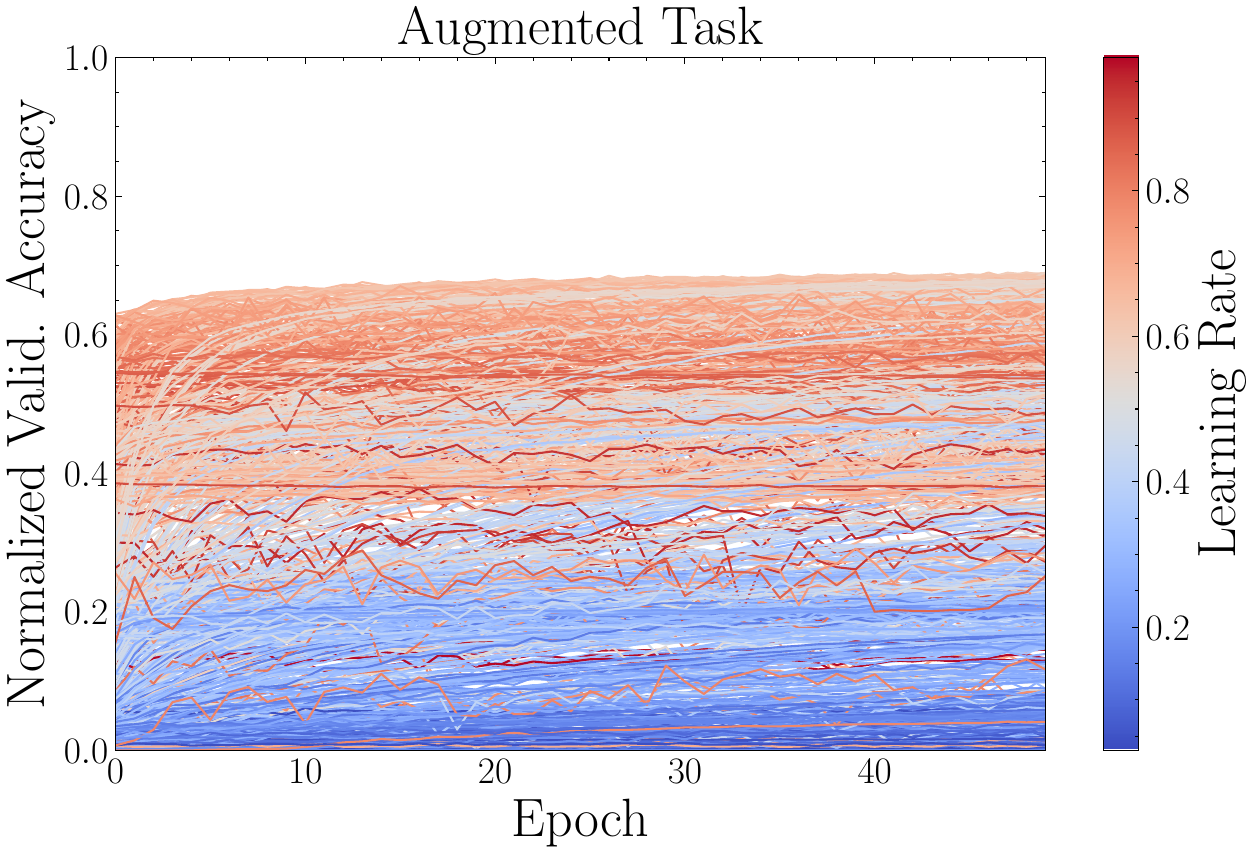}
            \caption{}
        \end{subfigure}
        \caption{(a) A selection of learning curves from TaskSet, the color of each curve is based on the learning rate value of the configuration. (b) The same set of learning curves after we apply Mixup learning curve augmentation.
        }
        \label{fig:mixup_augment}
    \end{figure}

    Figure \ref{fig:mixup_augment} shows the result of applying Mixup on the same set of learning curves as Figure \ref{fig:cdf_augment}. Due to the linear interpolation nature of Mixup, the resulting learning curves are smoother than the original, with some information, especially of the oscillating high learning curves, being phased out.

\section{Out-of-Distribution Results Per Task}
We considered the following tasks:
\begin{itemize}
    \item \textbf{Language Translation}: This task trains a transformer to perform language translation from German to English.
    \item \textbf{Word Language Model}: This task trains a neural network for language modeling by learning to predict the next word in a sequence. We evaluated two variants of this task, one using LSTMs \autocite{lstm_paper} and one using a transformer.
    \item \textbf{Sequence Predictor}: This task uses an LSTM to predict future values of sine wave signals.
    \item \textbf{GCN}: This task uses a Graph Convolutional Network \autocite{kipf_semi-supervised_2017} to perform node classification on the Cora dataset.
\end{itemize}
\label{appendix:ood_results_per_task}
    \begin{figure}[H]
        \centering
    
        \begin{minipage}[b]{0.48\linewidth}
            \centering
            \includegraphics[width=\linewidth]{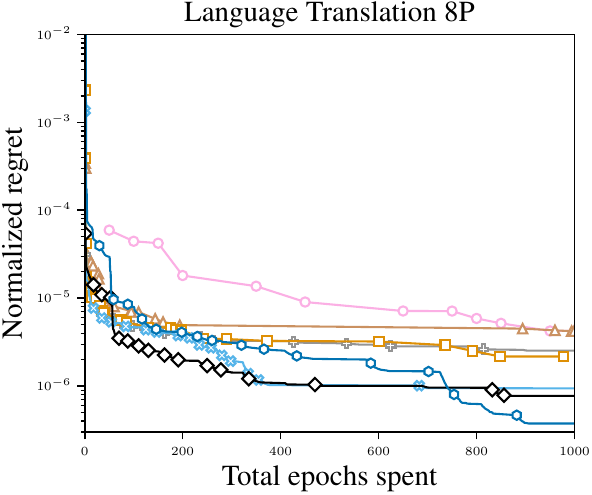}
        \end{minipage}
        \hfill
        \begin{minipage}[b]{0.48\linewidth}
            \centering
            \includegraphics[width=\linewidth]{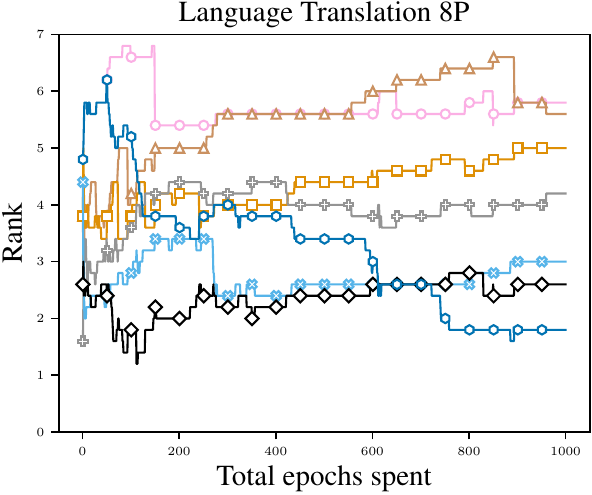}
        \end{minipage}
    
    
        \begin{minipage}[b]{0.48\linewidth}
            \centering
            \includegraphics[width=\linewidth]{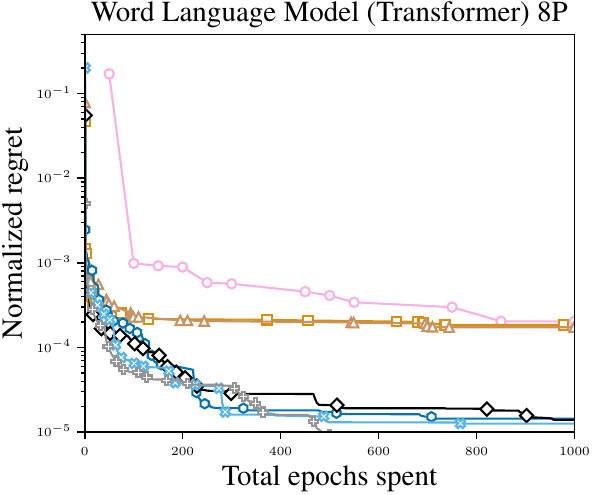}
        \end{minipage}
        \hfill
        \begin{minipage}[b]{0.48\linewidth}
            \centering
            \includegraphics[width=\linewidth]{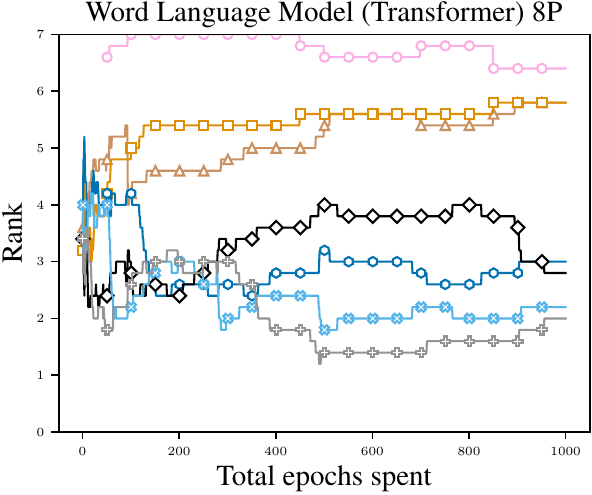}
        \end{minipage}
    
    
        \begin{minipage}[b]{0.8\linewidth}
            \centering
            \includegraphics[width=1.\linewidth]{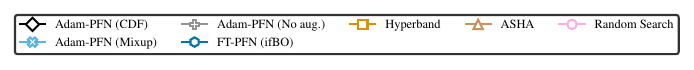}
            \label{fig:legend}
        \end{minipage}
    
        \caption{HPO results on real-world tasks.}
        \label{fig:pytorch_examples_hpo_appendix_1}
    \end{figure}

    \begin{figure}[H]
        \centering
    
        \begin{minipage}[b]{0.48\linewidth}
            \centering
            \includegraphics[width=\linewidth]{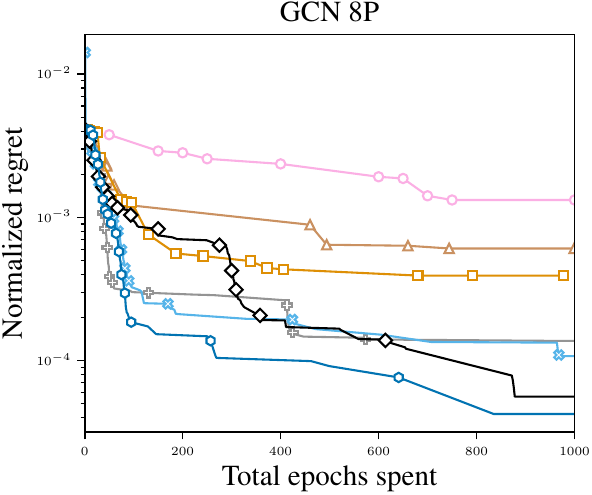}
        \end{minipage}
        \hfill
        \begin{minipage}[b]{0.48\linewidth}
            \centering
            \includegraphics[width=\linewidth]{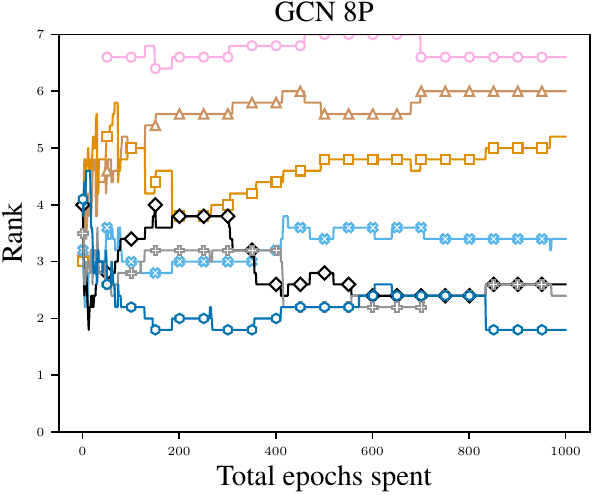}
        \end{minipage}
    
    
        \begin{minipage}[b]{0.48\linewidth}
            \centering
            \includegraphics[width=\linewidth]{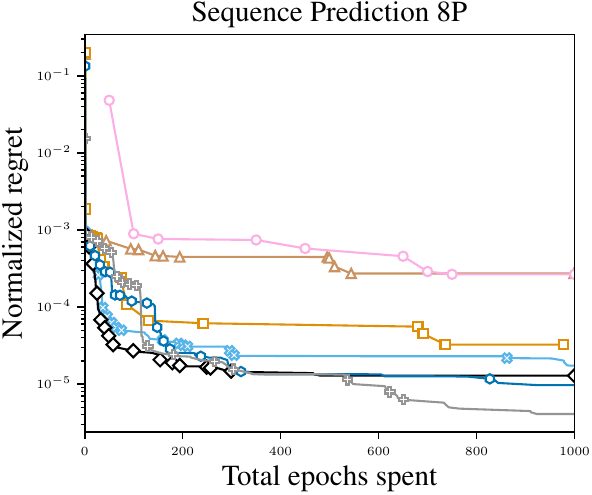}
        \end{minipage}
        \hfill
        \begin{minipage}[b]{0.48\linewidth}
            \centering
            \includegraphics[width=\linewidth]{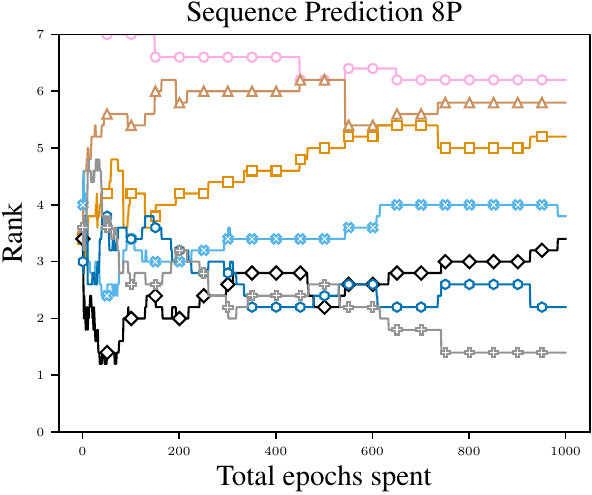}
        \end{minipage}
    
    
        \begin{minipage}[b]{0.48\linewidth}
            \centering
            \includegraphics[width=\linewidth]{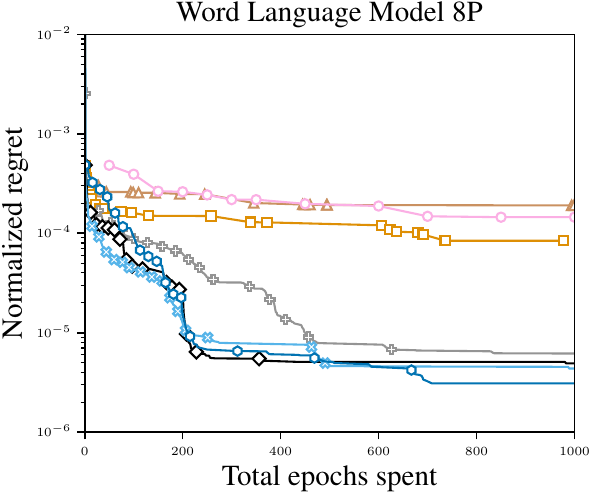}
        \end{minipage}
        \hfill
        \begin{minipage}[b]{0.48\linewidth}
            \centering
            \includegraphics[width=\linewidth]{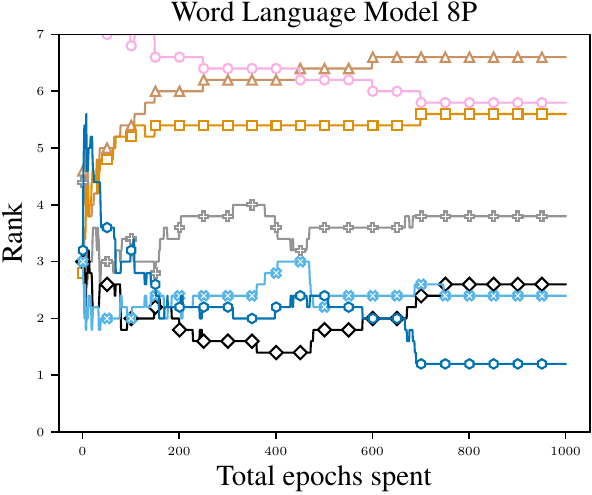}
        \end{minipage}
    
        \begin{minipage}[b]{0.8\linewidth}
            \centering
            \includegraphics[width=1.\linewidth]{images/paper/paper_gcn_8p_legend.pdf}
        \end{minipage}
    
        \caption{HPO results on real-world tasks.}
        \label{fig:pytorch_examples_hpo_appendix_2}
    \end{figure}

\section{Augmenting the HPs}
\label{appendix:hp_augmentation}
    Mixup augments both learning curves and HP configurations, so we also extended CDF-augment to support HP augmentation. Specifically, we transformed HP configurations using the cumulative distribution function (CDF) of the Beta distribution, similar to Equation \ref{eqn:cdf_y}.

    \begin{equation}
        x' = F_{Beta}(x;\mu, \kappa)
    \end{equation}

    To avoid extreme augmentation in HP space, we sample the concentration parameter $\kappa$ uniformly from the [2, 3] range, instead of [2, 5] that we used for learning curve augmentation. 

    To further isolate the effect of HP augmentation in Mixup, we performed an additional ablation study, in which we disabled the HP component and applied only Equation \ref{eqn::task_mixup_equation}. The learning curve extrapolation performance of these \texttt{Adam-PFN} variants is reported in Table \ref{tab:ablation_8p_horizontal} for different context sizes.

    Overall, the results indicate that HP augmentation negatively impacts performance. For the TaskSet-8P evaluation tasks (see Table \ref{tab:ablation_8p_horizontal}), the surrogate model performs worse when trained with HP augmentation for context sizes of 400 and 1000. The model performs equally well with the surrogate trained only with learning curve augmentation for a context size of 1600. HP augmentation also appears to hurt the performance of Mixup, since the results improved when we applied only task mixup. 
    
    \begin{table}[ht]
        \centering
        \small
        \scalebox{1.0}{
        \begin{tabular}{l ccc ccc ccc}
            \toprule
            \multicolumn{7}{c}{\textbf{TaskSet-8P}} \\
            \cmidrule(lr){2-7}
            \multirow{2}{*}{\textbf{Algorithm}} & 
            \multicolumn{2}{c}{\textbf{Context 400}} &
            \multicolumn{2}{c}{\textbf{Context 1000}} &
            \multicolumn{2}{c}{\textbf{Context 1600}} \\
            \cmidrule(lr){2-3} \cmidrule(lr){4-5} \cmidrule(lr){6-7}
            & LL & MSE & LL & MSE & LL & MSE \\
            \midrule
            \cdc CDF (w/ HP)           & \cdc5.309 & \cdc0.00055 & \cdc5.408 & \cdc0.00047 & \cdcb{5.441} & \cdcb{0.00043} \\
            Mixup (w/ HP)              & 4.916 & 0.00080 & 4.986 & 0.00067 & 5.022 & 0.00061 \\
            \cdc CDF   & \cdcb{5.326} & \cdcb{0.00054} & \cdcb{5.422} & \cdcb{0.00046} & \cdcb{5.441} & \cdcb{0.00043} \\
            Mixup (w/o HP)      & 5.296 & 0.00062 & 5.374 & 0.00051 & 5.409 & 0.00046 \\
            \cdc No Aug.       & \cdc4.947 & \cdc0.00062 & \cdc5.042 & \cdc0.00054 & \cdc5.066 & \cdc0.00052 \\
            \bottomrule
        \end{tabular}
        }
        \caption{Learning Curve extrapolation results for different augmentation methods. "w/o HP" indicates models trained without hyperparameter augmentation. "No Aug." is trained without any augmentation.}
        \label{tab:ablation_8p_horizontal}
    \end{table}

    We were somewhat cautious about these results, as the models may have overfitted to the HP configurations present on TaskSet, and thus may not be able to generalize to unseen configurations. To force the models to generalize during evaluation, we performed an additional experiment where we randomly withheld 400 HP configurations during surrogate model training and evaluated the performance exclusively on this held-out set for the evaluation tasks. The results are shown in Table \ref{tab:ablation_8p_lo_horizontal}.

    The results show that the models without HP configuration (\texttt{CDF} and \texttt{Mixup (w/o HP)}) were able to generalize to unseen HP configurations and also outperform their counterparts. Only learning curve augmentation seems to introduce enough variability in the data, and even though the transformer is trained on the same 600 HP configurations, it learns a sufficiently robust representation of the space, which allows generalization on previously unseen HPs.

    \begin{table}
        \centering
        \small
        \scalebox{1.0}{
        \begin{tabular}{l ccc ccc ccc}
            \toprule
            \multicolumn{7}{c}{\textbf{TaskSet-8P (400 Leave-Out)}} \\
            \cmidrule(lr){2-7}
            \multirow{2}{*}{\textbf{Algorithm}} & 
            \multicolumn{2}{c}{\textbf{Context 400}} & 
            \multicolumn{2}{c}{\textbf{Context 1000}} & 
            \multicolumn{2}{c}{\textbf{Context 1600}} \\
            \cmidrule(lr){2-3} \cmidrule(lr){4-5} \cmidrule(lr){6-7}
            & LL & MSE & LL & MSE & LL & MSE \\
            \midrule
            \cdc{CDF (w/ HP)}         & \cdc5.258 & \cdc0.00054 & \cdc5.380 & \cdc0.00045 & \cdc5.418 & \cdc0.00043 \\
            \cdc{CDF} & \cdcb{5.329} & \cdcb{0.00051} & \cdcb{5.418} & \cdcb{0.00042} & \cdcb{5.458} & \cdcb{0.00039} \\
            Mixup  (w/ HP)           & 4.850 & 0.00089 & 4.983 & 0.00078 & 5.035 & 0.00075 \\
            Mixup (w/o HP)     & 5.264 & 0.00053 & 5.357 & 0.00047 & 5.385 & 0.00044 \\
            \bottomrule
        \end{tabular}
        }
        \caption{Learning Curve extrapolation results for Adam-PFN variants with and without hyperparameter augmentation on 400 leave-out configurations.}
        \label{tab:ablation_8p_lo_horizontal}
    \end{table}

\end{document}